\documentclass[pdflatex,sn-mathphys-num]{sn-jnl}

\usepackage{graphicx}%
\usepackage{multirow}%
\usepackage{amsmath,amssymb,amsfonts}%
\usepackage{amsthm}%
\usepackage{mathrsfs}%
\usepackage[title]{appendix}%
\usepackage{xcolor}%
\usepackage{textcomp}%
\usepackage{manyfoot}%
\usepackage{booktabs}%
\usepackage{algorithm}%
\usepackage{algorithmicx}%
\usepackage{algpseudocode}%
\usepackage{listings}%
\usepackage{makecell}

\theoremstyle{thmstyleone}%

\theoremstyle{thmstyletwo}%

\theoremstyle{thmstylethree}%

\begin{document}

\title[Evolving CNN Architectures]{Evolving CNN Architectures: From Custom Designs to Deep Residual Models for Diverse Image Classification and Detection Tasks}

\author[1]{\fnm{Mahmudul} \sur{Hasan}}\email{mahmudul.hhh@gmail.com\textsuperscript{1}}
\equalcont{These authors contributed equally to this work.}

\author[2]{\fnm{Mabsur Fatin Bin} \sur{Hossain}}\email{mansifhossain5@gmail.com\textsuperscript{2}}
\equalcont{These authors contributed equally to this work.}

\affil[]{
\orgdiv{Department of Computer Science and Engineering},
\orgname{University of Dhaka}, 
\orgaddress{\street{University Street}, \city{Dhaka}, \postcode{1000}, \country{Bangladesh}}
}

\abstract{
This paper presents a comparative study of a custom convolutional neural network (CNN) architecture against widely used pretrained and transfer learning CNN models across five real-world image datasets. The datasets span binary classification, fine-grained multiclass recognition, and object detection scenarios. We analyze how architectural factors, such as network depth, residual connections, and feature extraction strategies, influence classification and localization performance. The results show that deeper CNN architectures provide substantial performance gains on fine-grained multiclass datasets, while lightweight pretrained and transfer learning models remain highly effective for simpler binary classification tasks. Additionally, we extend the proposed architecture to an object detection setting, demonstrating its adaptability in identifying unauthorized auto-rickshaws in real-world traffic scenes. Building upon a systematic analysis of custom CNN architectures alongside pretrained and transfer learning models, this study provides practical guidance for selecting suitable network designs based on task complexity and resource constraints.
}

\maketitle

\section{Introduction}

Image classification has become one of the most widely adopted applications of deep learning, powering real-world systems in urban monitoring, agriculture, environmental assessment, and automated quality control. Convolutional Neural Networks (CNNs) have played a central role in these advances due to their ability to learn hierarchical visual features directly from raw images. Despite the strong performance of well-established architectures, designing compact and task‑specific CNN models remains important, particularly when datasets vary widely in scale, domain complexity, and visual distributions.

In this study, we evaluate the performance of CNN-based models across five diverse image datasets spanning both binary and multiclass settings. These datasets cover practical real-world challenges: road surface analysis, footpath encroachment detection, fruit variety recognition, and paddy species classification. The \emph{Road Damage}\cite{hossen2025roaddamagemanholedetection} and \emph{FootpathVision}\cite{Lubaina2025FootpathVision} datasets represent binary classification tasks focused on urban infrastructure monitoring, where images capture conditions such as damaged versus undamaged roads or encroached versus clear sidewalks. In contrast, the \emph{MangoImageBD}\cite{FERDAUS2025111908} and \emph{PaddyVarietyBD}\cite{TAHSIN2025111514} datasets present large‑scale multiclass problems involving the identification of multiple mango varieties and microscopic paddy kernels, respectively. Additionally, we include an \emph{Auto-Rickshaw Detection}\cite{sukanto2025detectingunauthorizedvehiclesusing} dataset that introduces an object-recognition challenge, where the goal is to differentiate motorized auto-rickshaws from visually similar non-motorized rickshaws in complex traffic scenes. This dataset adds a spatial localization component and provides insight into how classification-oriented CNN architectures adapt to images containing multiple overlapping objects. Together, these datasets provide a comprehensive evaluation environment with varying image resolutions, visual characteristics, and class distributions.

Given this diversity, relying solely on standard deep architectures may not yield optimal performance or adaptability across tasks. This motivates the exploration of custom CNN designs tailored to specific feature patterns and dataset characteristics. In this work, we investigate a custom convolutional architecture and compare it against progressively simplified variants and an evolved baseline model to understand how different architectural choices influence performance across multiple domains. Our goal is to provide a systematic analysis of how modifications in convolutional structure, residual connections, and feature extraction depth affect classification accuracy on heterogeneous real-world datasets. To complement this analysis, we evaluate two widely used pretrained CNN architectures---MobileNet and EfficientNet---under both pretrained and transfer learning setups, and compare their performance against the best-performing variants of our custom CNN across multiple image classification tasks. For the object detection dataset, we adopt state-of-the-art detection frameworks, namely YOLO and Faster R-CNN, to assess localization and recognition performance in complex traffic scenes.

\section{Proposed Custom CNN Architecture}

In this section, we introduce a custom convolutional neural network designed to efficiently extract hierarchical visual features across diverse image classification tasks. Our goal was to balance model complexity and representational power, enabling robust recognition across datasets with varying visual characteristics while keeping the parameter count manageable for practical deployment.

The architecture is organized into modular components that progressively transform raw image inputs into high-level semantic representations. We begin with a compact convolutional feature extractor to capture low-level spatial patterns, followed by a series of residual blocks incorporating depthwise separable convolutions for computational efficiency. Finally, a classification head aggregates global information and produces the final predictions.

This design allows the network to learn multi-scale features, handle complex spatial dependencies, and maintain robustness across different object categories, visual domains, and environmental conditions. Each component of the architecture is described in detail below.

\subsection{Initial Feature Extractor: Stacked 3$\times$3 Convolutional Layers}

To initiate feature extraction, our custom CNN begins with three stacked 3$\times$3 convolutional layers. This design is inspired by the VGG family of networks, which demonstrated that stacking smaller convolutional kernels (e.g., 3$\times$3) can be more effective and parameter-efficient than using a single large kernel. The use of multiple nonlinearities between layers also allows the model to learn more complex local patterns.

The configuration of the initial block is as follows:
\begin{itemize}
    \item The first layer is a 3$\times$3 convolution with 32 filters and a stride of 2, followed by Batch Normalization and ReLU activation. The stride reduces the spatial resolution while preserving important low-level features.
    \item The second layer is another 3$\times$3 convolution with 32 filters, using a stride of 1 and followed by Batch Normalization and ReLU activation.
    \item The third layer is a 3$\times$3 convolution with 64 filters, again followed by Batch Normalization and ReLU activation.
    \item A \textbf{MaxPooling2D} layer with a pool size of 3$\times$3 and a stride of 2 is applied at the end to further downsample the feature maps and retain the most salient features.
\end{itemize}

This initial convolutional stack acts as a compact and efficient feature extractor, enabling the model to capture localized spatial patterns early in the network.

\subsection{Residual Block Design with Depthwise Separable Convolutions}

A central component of our architecture is the custom residual block, which utilizes \textit{depthwise separable convolutions} to reduce parameter count and computational overhead. This design draws inspiration from MobileNet and replaces traditional convolutional layers with a combination of \textbf{DepthwiseConv2D} and \textbf{Pointwise Conv2D} (1$\times$1 convolution).

Unlike standard 3$\times$3 convolutions that operate across all channels simultaneously, a \textbf{depthwise convolution} applies one filter per input channel independently. This enables spatial feature extraction without channel mixing. A subsequent 1$\times$1 pointwise convolution then recombines the output across channels, allowing for cross-channel interactions.

The structure of the residual block is as follows:
\begin{itemize}
    \item A \textbf{DepthwiseConv2D} layer with a 3$\times$3 kernel and a configurable stride, followed by Batch Normalization.
    \item A \textbf{Pointwise Conv2D} (1$\times$1) to mix channel information, followed by Batch Normalization and ReLU activation.
    \item A second \textbf{3$\times$3 Conv2D} layer to enhance feature representation, followed by Batch Normalization.
    \item A \textbf{shortcut connection}, which uses a 1$\times$1 projection if the input and output dimensions differ.
    \item Final element-wise addition and ReLU activation to complete the residual connection.
\end{itemize}

\subsection{Residual Block Stacking in the Full Architecture}

After the initial stack of 3$\times$3 convolutional layers, we organize the rest of the architecture into four sequential stages, each comprising two custom residual blocks as previously defined. The number of filters increases progressively across stages to allow the model to learn increasingly abstract and high-dimensional features, while spatial resolution is reduced through strided convolutions at the beginning of each stages.

The overall block-wise configuration is as follows:
\begin{itemize}
    \item \textbf{Stage 1:} Two residual blocks with 64 filters, both with stride 1.
    \item \textbf{Stage 2:} Two residual blocks with 128 filters. The first uses stride 2 for downsampling; the second uses stride 1.
    \item \textbf{Stage 3:} Two residual blocks with 256 filters. The first uses stride 2; the second uses stride 1.
    \item \textbf{Stage 4:} Two residual blocks with 512 filters. The first uses stride 2; the second uses stride 1.
\end{itemize}

This hierarchical stacking strategy enables the network to extract and refine features at multiple spatial resolutions, while retaining efficiency through depthwise separable convolutions.

\subsection{Classification Head}

After the final stage of residual blocks, the network transitions from spatial feature maps to a compact representation suitable for classification. To achieve this, we apply a \textbf{Global Average Pooling} layer, which aggregates each feature map into a single value by averaging over its spatial dimensions. This reduces the parameter count significantly compared to flattening while preserving the most salient global features.

Following the pooling layer, we include a fully connected layer with 128 units and ReLU activation. This layer introduces an additional level of non-linearity and enables the model to combine high-level features extracted by the convolutional backbone. This dense layer was intentionally added as a design variation to enhance representational capacity.

The final output layer is a Dense layer with two units and a softmax activation function, suitable for binary classification tasks. The complete classification head is summarized as follows:
\begin{itemize}
    \item \textbf{GlobalAveragePooling2D} to convert feature maps into a global descriptor.
    \item \textbf{Dense(128)} with ReLU activation to provide an additional learned feature transformation.
    \item \textbf{Dense(2)} with softmax activation for binary classification output.
\end{itemize}

\section{Intermediate Variants and Evolution}

To better understand and explore our architectural choices, we investigate progressively evolved variants of our custom CNN architecture. Each variant introduces specific changes to components of the original design, enabling us to analyze the effect of individual architectural decisions on model performance.

\subsection{Variant A: Standard Residual Blocks}

This variant retains the same initial stack of 3$\times$3 convolutional layers and the fully connected classification head used in our custom model. However, it replaces the residual blocks with a conventional two-layer convolutional structure.

All other architectural components—including the number of filters, the four-stage residual stacking strategy, and the use of global average pooling—remain unchanged.

\subsection{Variant B: Initial 7$\times$7 Convolution}

In this variant, we replace the initial stack of three 3$\times$3 convolutional layers with a single 7$\times$7 convolutional layer with stride 2. This configuration offers a broader receptive field at the very first layer, allowing the model to capture coarser spatial patterns early in the network.

The rest of the architecture remains the same as in Variant A, including the use of standard convolutional residual blocks, global average pooling, and the fully connected classification head. This design explores the trade-off between early-layer depth and spatial abstraction, helping us understand the impact of reducing initial non-linearity in favor of larger spatial coverage.

\subsection{Variant C: Evolved Baseline Model}

This configuration represents the most performant version among all evolved variants in our binary classification experiments. It builds upon Variant B by removing the additional fully connected layer before the output, thereby simplifying the classification head.

The overall structure includes:
\begin{itemize}
    \item A single 7$\times$7 convolutional layer for initial feature extraction.
    \item Standard convolutional residual blocks organized in four stages, as described previously.
    \item Global average pooling directly followed by a final Dense layer with softmax activation for binary classification.
\end{itemize}

We refer to this model as our \textit{evolved baseline} due to its balanced design and superior performance across datasets. It serves as a reference point for evaluating other variants in subsequent sections.

\subsection{Enhanced Baseline with Bottleneck Residual Blocks}

To handle more complex multiclass classification tasks, we extend our evolved baseline by incorporating \textit{bottleneck residual blocks}—a design known for improving computational efficiency and gradient flow in deeper networks. These blocks reduce the number of parameters while preserving representational power, making them suitable for high-resolution image recognition.

Each bottleneck residual block consists of three convolutional layers: a 1$\times$1 convolution for dimensionality reduction, a 3$\times$3 convolution for spatial feature extraction, and another 1$\times$1 convolution to restore the channel dimension. A projection shortcut is applied whenever the input and output dimensions do not match. The overall structure of the network includes four stages, each comprising multiple such bottleneck blocks with increasing depth:

\begin{itemize}
    \item \textbf{Stage 1}: 3 bottleneck residual blocks with output dimensions of 256. The first block includes a projection shortcut even though the stride is 1.
    \item \textbf{Stage 2}: 4 bottleneck residual blocks with output dimensions of 512. The first block uses a stride of 2 for spatial downsampling.
    \item \textbf{Stage 3}: 6 bottleneck residual blocks with output dimensions of 1024. Downsampling is again performed in the first block using stride 2.
    \item \textbf{Stage 4}: 3 bottleneck residual blocks with output dimensions of 2048, beginning with a downsampling block.
\end{itemize}

This deeper architecture allows the model to better capture hierarchical features necessary for distinguishing between a larger number of classes in more diverse datasets. All stages are followed by a global average pooling layer and a dense classification head appropriate for the number of output classes.

\subsection{Adaptation for Object-Level Detection}

For the object-level detection task involving auto-rickshaws in traffic scenes, we adopt a
lightweight, custom-designed detection model inspired by the core principles of single-stage detectors such as YOLO, implemented in a simplified form tailored to our dataset and
experimental scope. Unlike our classification-oriented CNN architectures, this model is explicitly designed to jointly predict object category labels and bounding box coordinates within a unified framework. 

The proposed detection network, referred to as \textit{MiniYOLO}, consists of a compact convolutional backbone followed by a unified prediction head. The backbone comprises a sequence of convolutional layers with gradually increasing channel depth (16, 32, 64, and 128 filters), each using $3\times3$ kernels and LeakyReLU activations. Max-pooling layers are applied after early convolutional blocks to progressively reduce spatial resolution while retaining salient visual features. An adaptive average pooling layer is used at the end of the feature extractor to produce a fixed-dimensional representation irrespective of input image size.

The detection head is implemented as a fully connected layer that outputs a combined prediction vector consisting of class logits and bounding box parameters. Specifically, for each input image, the model predicts:
\begin{itemize}
    \item A set of class scores corresponding to the predefined object categories.
    \item Four continuous values representing the bounding box coordinates.
\end{itemize}

During training, the model optimizes a composite loss function that combines a classification loss and a localization loss. Cross-entropy loss is used for object class prediction, while mean squared error (MSE) loss is applied to bounding box regression. Localization loss is computed only for samples containing valid object annotations, ensuring that background images do not introduce spurious regression errors.

This lightweight, single-stage design allows the model to perform object recognition and localization efficiently, making it suitable for scenarios with limited computational resources. While simpler than full-scale detection frameworks, the MiniYOLO model provides a practical baseline for evaluating object-level detection performance and enables direct comparison with more advanced detectors such as YOLO and Faster R-CNN in subsequent sections.

\section{Pretrained CNN Models and Transfer Learning Approach}

We divide our analysis into two parts: one for image classification tasks using MobileNetV2 and EfficientNetB0 under both pretrained and transfer learning setups, and another for object detection tasks using state-of-the-art detection frameworks such as YOLO and Faster R-CNN, also evaluated under pretrained and transfer learning strategies.

\subsection{Pretrained and Transfer Learning Models for Classification Tasks}

To enable a fair comparison with our custom CNN trained from scratch, we evaluate two widely used CNN architectures---MobileNetV2 and EfficientNetB0---initialized with ImageNet pretrained weights. Overall, our experiments are organized into three distinct model categories:

\begin{itemize}
    \item \textbf{Custom CNN models}, trained from scratch with randomly initialized weights.
    \item \textbf{Pretrained models}, where popular CNN architectures are initialized with ImageNet weights and trained end-to-end on the target datasets.
    \item \textbf{Transfer learning models}, where the pretrained convolutional backbone is frozen and only task-specific classification layers are fine-tuned.
\end{itemize}

For the transfer learning models, we retain the pretrained convolutional backbone as a fixed feature extractor by setting \texttt{base\_model.trainable = False}. To adapt the network to each classification task and label set, we append a custom classification head consisting of a \textbf{GlobalAveragePooling2D} layer for spatial dimensionality reduction, a \textbf{Dropout} layer for regularization, and a \textbf{Dense} output layer with softmax activation, where the number of output units matches the number of classes in the target dataset.

The classification head is trained from scratch, enabling the model to learn dataset-specific label distributions while benefiting from the general visual representations encoded in the frozen backbone.

\subsection{Pretrained and Transfer Learning Models for Object Detection Tasks}

To evaluate object-level detection performance, we employ two widely used detection frameworks—Faster R-CNN and YOLO—initialized with pretrained weights. Similar to the classification experiments, our object detection analysis is conducted under both pretrained and transfer learning settings.

For Faster R-CNN, models with ResNet-50-FPN and MobileNetV3-FPN backbones are initialized using pretrained weights. To adapt these models to the auto-rickshaw detection task, the original detection head is replaced with a task-specific prediction head matching the number of target classes. In the transfer learning setup, the backbone network is frozen and only the region proposal and detection heads are trained. In the fine-tuning phase, all layers are unfrozen and jointly optimized using a reduced learning rate.

For YOLO, we adopt a pretrained YOLOv8 model and follow a two-stage training strategy. In the transfer learning phase, the backbone layers are frozen and only the detection heads are trained on the target dataset. This is followed by a fine-tuning phase in which the entire network is unfrozen and trained end-to-end with a lower learning rate to further adapt the model to the traffic scene domain.

\section{Results}

In this section, we present the evaluation outcomes of our models across all datasets. We report standard classification metrics, including accuracy, precision, recall, and F1-score,  training time, and model size.

\subsection{Internal Architecture Comparisons (Custom Models Only)}

We begin by evaluating multiple variants of our custom CNN architecture to identify the most effective internal design for each dataset. These results establish baselines for subsequent comparisons with pretrained and transfer learning models.

\subsubsection{Binary Classification: Road Damage Dataset}

Table~\ref{tab:road_damage_results} shows the classification performance of the four model variants—Custom CNN, Variant A, Variant B, and Evolved Baseline—on the binary Road Damage classification task.

\begin{table}[!t]
\caption{Performance on Road Damage Dataset (Test Set)}
\label{tab:road_damage_results}
\centering
\begin{tabular}{@{}lcccc@{}}
\toprule
\textbf{Model} & \textbf{Accuracy} & \textbf{Precision} & \textbf{Recall} & \textbf{F1-Score} \\
\midrule
Custom CNN         & 0.72 & 0.52 & 0.72 & 0.60 \\
Variant A          & 0.72 & 0.52 & 0.72 & 0.60 \\
Variant B          & 0.31 & 0.80 & 0.31 & 0.18 \\
Evolved Baseline   & 0.78 & 0.77 & 0.78 & 0.75 \\
\bottomrule
\end{tabular}
\end{table}

\begin{figure}[H]
\centering

% Row 1
\begin{minipage}[b]{0.48\textwidth}
    \centering
    \includegraphics[width=0.95\linewidth]{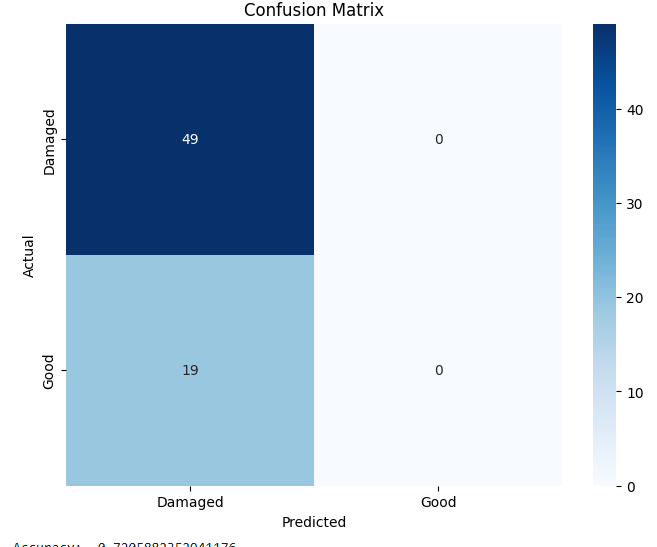}
    \par\vspace{1ex}(a) Custom CNN
\end{minipage}
\hfill
\begin{minipage}[b]{0.48\textwidth}
    \centering
    \includegraphics[width=0.95\linewidth]{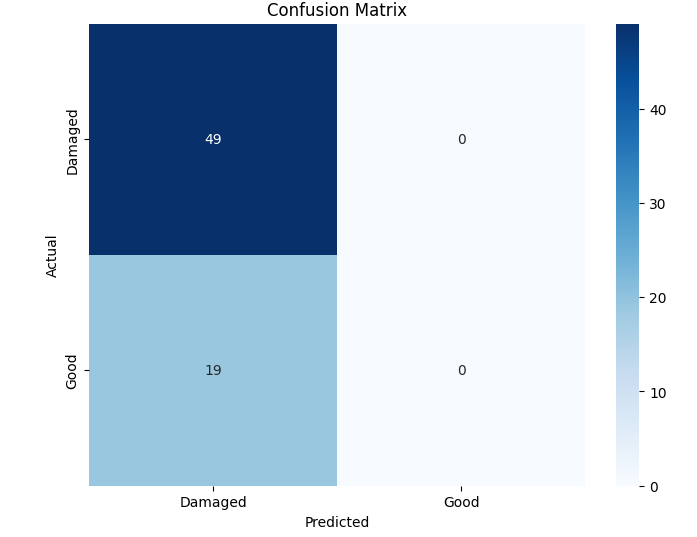}
    \par\vspace{1ex}(b) Variant A
\end{minipage}

\vspace{0.5cm}

% Row 2
\begin{minipage}[b]{0.48\textwidth}
    \centering
    \includegraphics[width=0.95\linewidth]{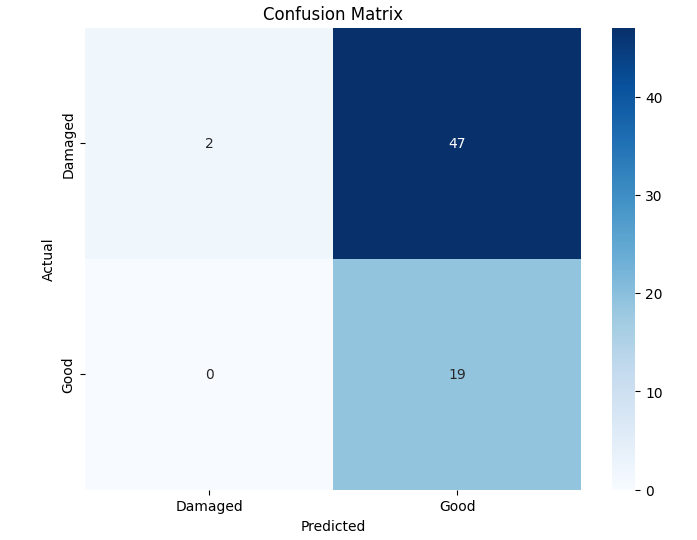}
    \par\vspace{1ex}(c) Variant B
\end{minipage}
\hfill
\begin{minipage}[b]{0.48\textwidth}
    \centering
    \includegraphics[width=0.95\linewidth]{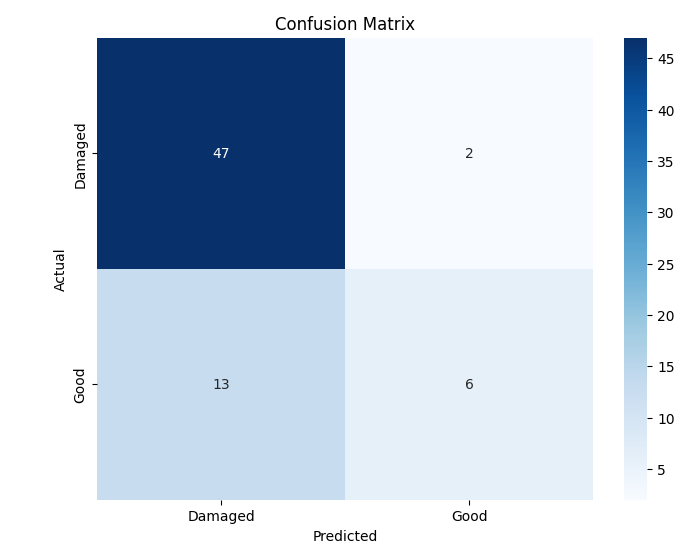}
    \par\vspace{1ex}(d) Evolved Baseline
\end{minipage}

\caption{Confusion matrices of four models on the Road Damage Dataset (Test Set).}
\label{fig:road_damage_confusions}
\end{figure}

Among the four models, the Evolved Baseline demonstrated the highest overall performance across all metrics, indicating robust generalization to the test data. Notably, it successfully predicted both classes—damaged and undamaged roads—even though the Good Roads class had fewer training samples. In contrast, the Custom CNN and Variant A, while achieving moderate accuracy and recall, largely failed to predict the minority class and focused almost exclusively on damaged roads. This suggests that the Evolved Baseline was more resilient to class imbalance, a desirable trait in real-world deployments where certain categories may be underrepresented. Variant B, while showing high precision, suffered from extremely low recall, implying a tendency to misclassify most samples as one class. These class-level behaviors are further evident in the confusion matrices. Figure~\ref{fig:road_damage_confusions} presents the confusion matrices for all four model variants, illustrating their class-wise prediction behavior on the Road Damage test set.

\subsubsection{Binary Classification: Footpath Dataset}

Table~\ref{tab:footpath_results} shows the classification performance of the four model variants—Custom CNN, Variant A, Variant B, and Evolved Baseline—on the binary Footpath classification task.

Among the four models, the \textbf{Evolved Baseline} achieved the strongest performance, reaching an accuracy of \textbf{0.71} and balanced precision, recall, and F1-scores of \textbf{0.70}. This indicates that it was able to effectively distinguish between encroached and unencroached footpaths despite the challenging variations present in the dataset. \textbf{Variant~B} also demonstrated reasonable performance, showing consistent metrics across all evaluation measures (0.61), suggesting that its architecture handled the dataset moderately well.

In contrast, both the \textbf{Custom CNN} and \textbf{Variant~A} struggled to identify the encroached class reliably. The Custom CNN, in particular, predicted almost all samples as unencroached, leading to poor precision and F1-score for the minority class. Variant~A showed high precision but extremely low recall for encroached regions, reflecting a tendency to overpredict the majority class.

\begin{figure}[!t]
\centering

% Row 1
\begin{minipage}[b]{0.48\textwidth}
    \centering
    \includegraphics[width=0.95\linewidth]{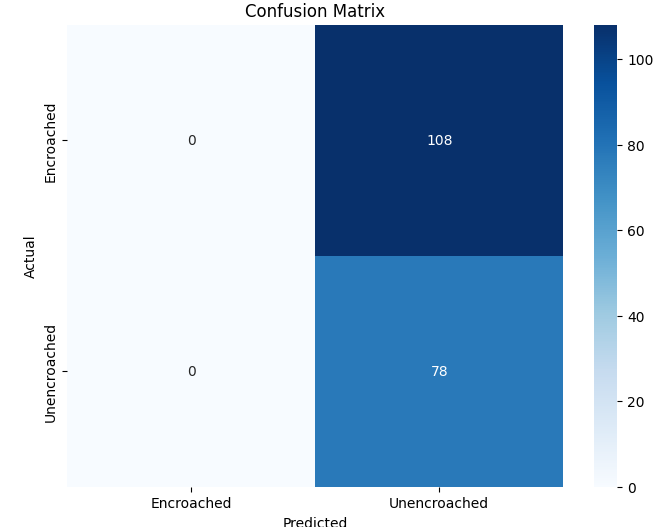}
    \par\vspace{1ex}(a) Custom CNN
\end{minipage}
\hfill
\begin{minipage}[b]{0.48\textwidth}
    \centering
    \includegraphics[width=0.95\linewidth]{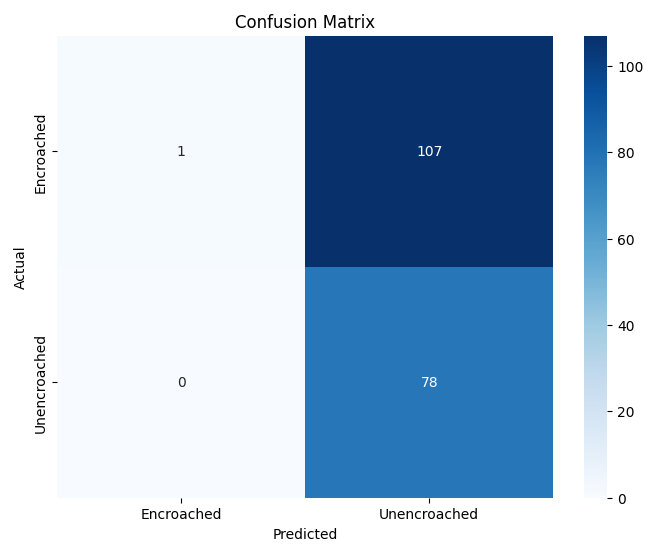}
    \par\vspace{1ex}(b) Variant A
\end{minipage}

\vspace{0.5cm}

% Row 2
\begin{minipage}[b]{0.48\textwidth}
    \centering
    \includegraphics[width=0.95\linewidth]{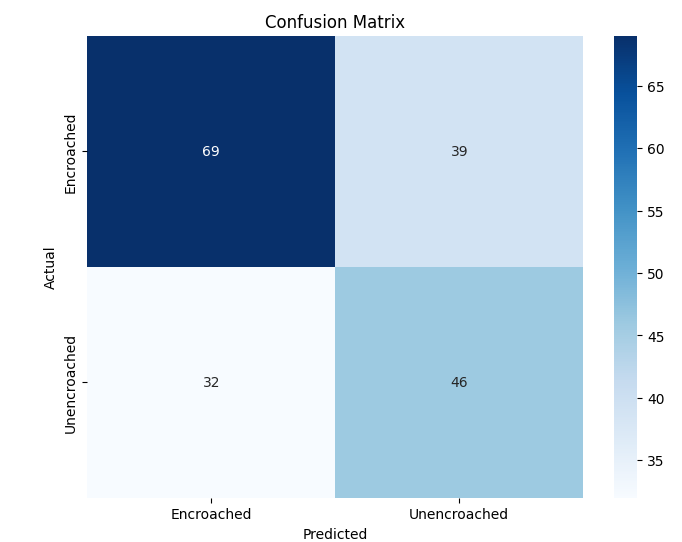}
    \par\vspace{1ex}(c) Variant B
\end{minipage}
\hfill
\begin{minipage}[b]{0.48\textwidth}
    \centering
    \includegraphics[width=0.95\linewidth]{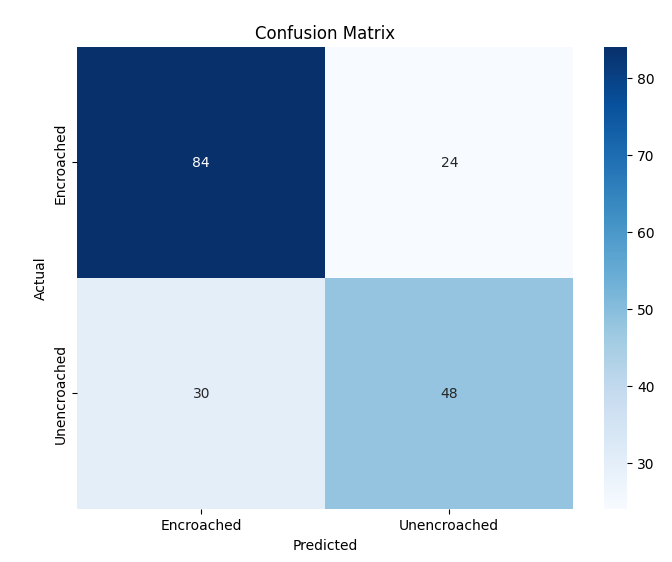}
    \par\vspace{1ex}(d) Evolved Baseline
\end{minipage}

\caption{Confusion matrices of four models on the Footpath Dataset (Test Set).}
\label{fig:footpath_confusions}
\end{figure}

The confusion matrices in Figure~\ref{fig:footpath_confusions} further illustrate these differences, showing how each model distributes predictions across the two classes.

\begin{table}[!t]
\caption{Performance on Footpath Dataset (Test Set)}
\label{tab:footpath_results}
\centering
\begin{tabular}{@{}lcccc@{}}
\toprule
\textbf{Model} & \textbf{Accuracy} & \textbf{Precision} & \textbf{Recall} & \textbf{F1-Score} \\
\midrule
Custom CNN         & 0.42 & 0.21 & 0.50 & 0.30 \\
Variant A          & 0.42 & 0.71 & 0.50 & 0.31 \\
Variant B          & 0.62 & 0.61 & 0.61 & 0.61 \\
Evolved Baseline   & 0.71 & 0.70 & 0.70 & 0.70 \\
\bottomrule
\end{tabular}
\end{table}

\subsubsection{Multiclass Classification: MangoImageBD}

The \textbf{MangoImageBD} dataset presents a fine-grained multiclass classification challenge involving 15 visually similar mango varieties. These categories differ subtly in colour tone, surface texture, and shape—making this task significantly more difficult than the binary datasets discussed earlier. To handle this increased complexity, we employ a \textit{deeper baseline model} incorporating \textbf{bottleneck residual blocks} (Section~3.4), which expands representational capacity while maintaining computational efficiency.

We first evaluated the performance of the \emph{baseline model} (without bottleneck blocks) to establish a reference point. The classification report for this model indicated relatively poor performance, with an overall accuracy of approximately 0.35 and low precision and F1-scores. This suggests that the baseline design, while effective for binary datasets, lacked the necessary depth and feature abstraction required for fine-grained multiclass recognition.

Next, we trained the \emph{deeper baseline model} (with bottleneck blocks) on the same dataset. As shown in the detailed classification report, this deeper architecture achieved a substantial performance boost, reaching an accuracy of \textbf{0.98}, with similarly high precision, recall, and F1-score. This improvement demonstrates the importance of increased depth and channel expansion when dealing with multiclass agricultural datasets, where inter-class boundaries are visually subtle.

We now present the performance of both models in Table~\ref{tab:mango_results}.

\begin{table}[!t]
\caption{Performance on MangoImageBD Dataset (Test Set)}
\label{tab:mango_results}
\centering
\begin{tabular}{@{}lcccc@{}}
\toprule
\textbf{Model} & \textbf{Accuracy} & \textbf{Precision} & \textbf{Recall} & \textbf{F1-Score} \\
\midrule
Baseline Model (No Bottleneck) & 0.35 & 0.32 & 0.35 & 0.26 \\
Deeper Baseline Model (With Bottleneck) & 0.98 & 0.98 & 0.98 & 0.98 \\
\bottomrule
\end{tabular}
\end{table}

\begin{figure}[!t]
\centering

% Row 1
\begin{minipage}[b]{0.7\textwidth}
    \centering
    \includegraphics[width=\linewidth]{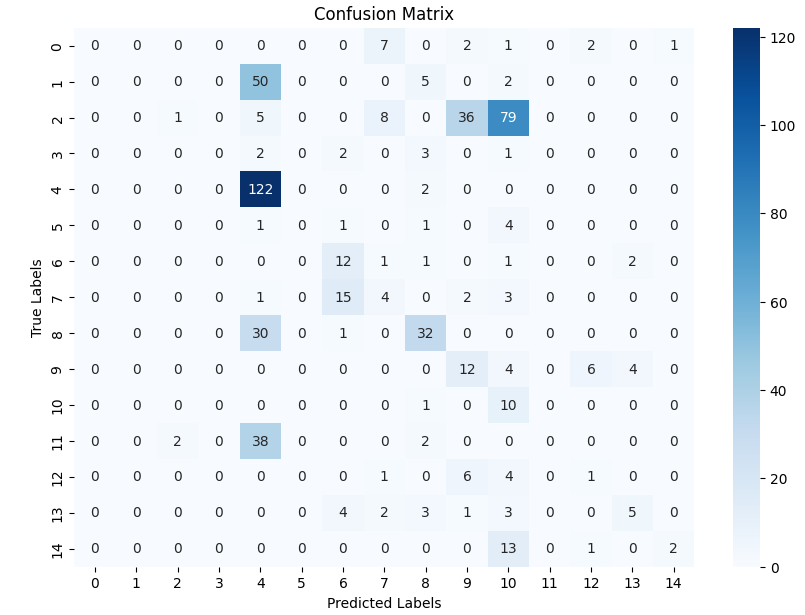}
    \par\vspace{1ex}(a) Baseline Model (No Bottleneck)
\end{minipage}

\vspace{0.8cm}

% Row 2
\begin{minipage}[b]{0.8\textwidth}
    \centering
    \includegraphics[width=\linewidth]{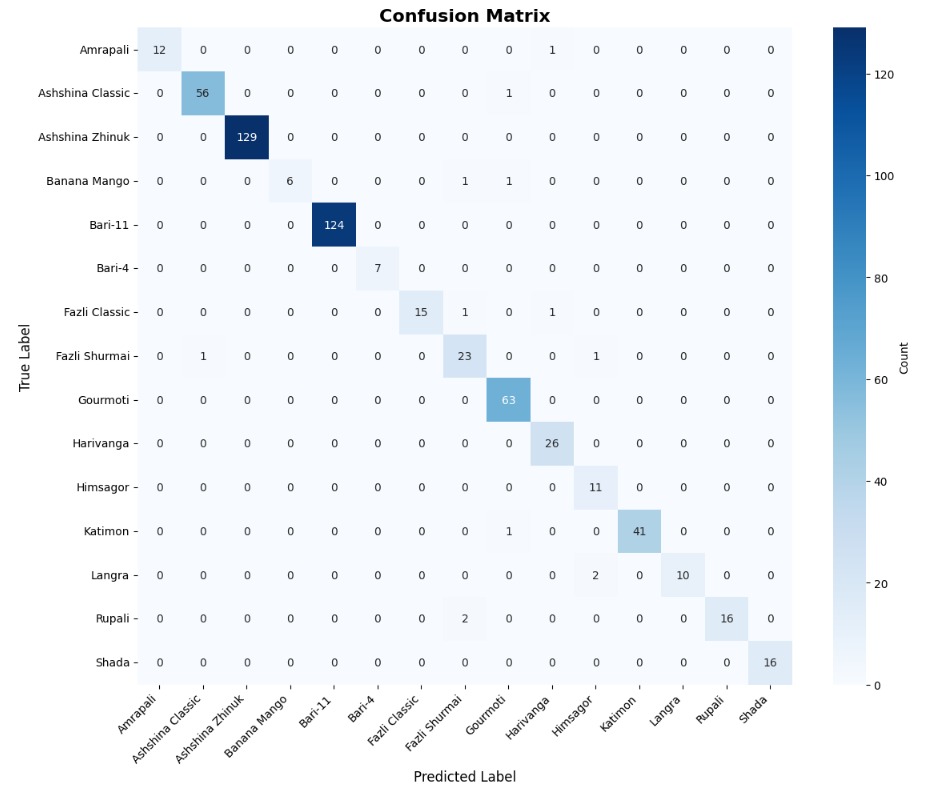}
    \par\vspace{1ex}(b) Deeper Baseline Model (With Bottleneck)
\end{minipage}

\caption{Confusion matrices of the two models on the MangoImageBD Dataset (Test Set).}
\label{fig:mango_confusions}
\end{figure}

Then, we also present the confusion matrices of both models in Figure~\ref{fig:mango_confusions}, which illustrate their class-wise prediction behavior on the MangoImageBD test set.

\subsubsection{Multiclass Classification: PaddyVarietyBD}

The \textbf{PaddyVarietyBD} dataset presents another challenging multiclass classification task, involving microscopic images of 35 paddy grain types. Compared to MangoImageBD, this dataset features higher inter-class visual similarity and lower-resolution texture details, making it particularly demanding for deep learning models. To handle this fine-grained discrimination task, we apply the same architectural setup used in the previous section—comparing a \textit{baseline model} (without bottleneck blocks) against a \textit{deeper baseline model} (with bottleneck residual blocks; see Section~3.4).

The \emph{baseline model} struggled to generalize effectively on this dataset. As shown in the classification metrics, it achieved an accuracy of just \textbf{0.43}, with relatively low precision and F1-score. These results suggest that the model failed to extract discriminative high-level features, likely due to its limited depth and representational power.

To improve performance, we employed the \emph{deeper baseline model} with stacked bottleneck residual blocks. This version yielded significant gains, achieving an accuracy of \textbf{0.71}, with balanced precision and F1-score. These improvements indicate that increasing the network depth and feature dimensionality helped the model better capture the subtle variations across paddy categories.

\begin{figure}[H]
\centering

% Row 1 – Baseline Model
\begin{minipage}[b]{0.85\textwidth}
    \centering
    \includegraphics[width=\linewidth]{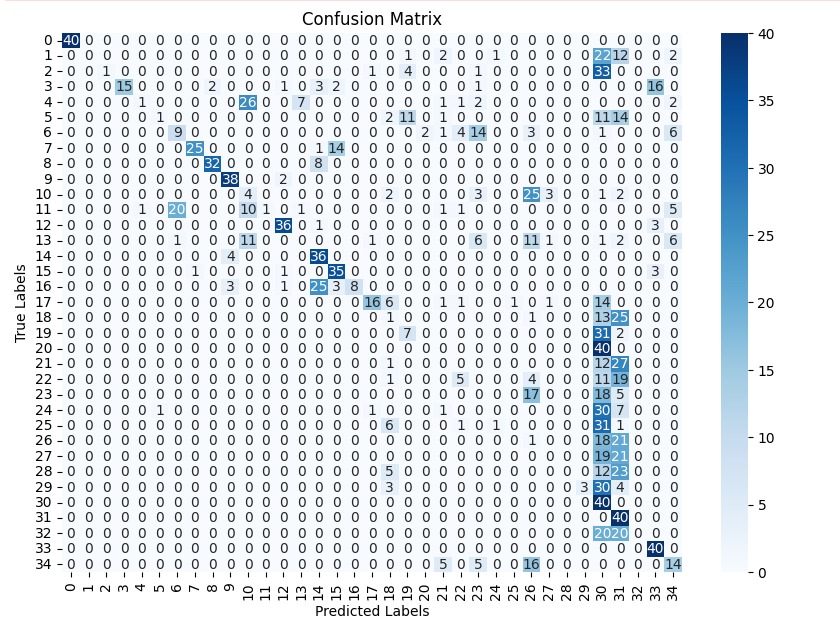}
    \par\vspace{1ex}(a) Baseline Model (No Bottleneck)
\end{minipage}

\vspace{0.8cm}

% Row 2 – Deeper Baseline Model
\begin{minipage}[b]{0.85\textwidth}
    \centering
    \includegraphics[width=\linewidth]{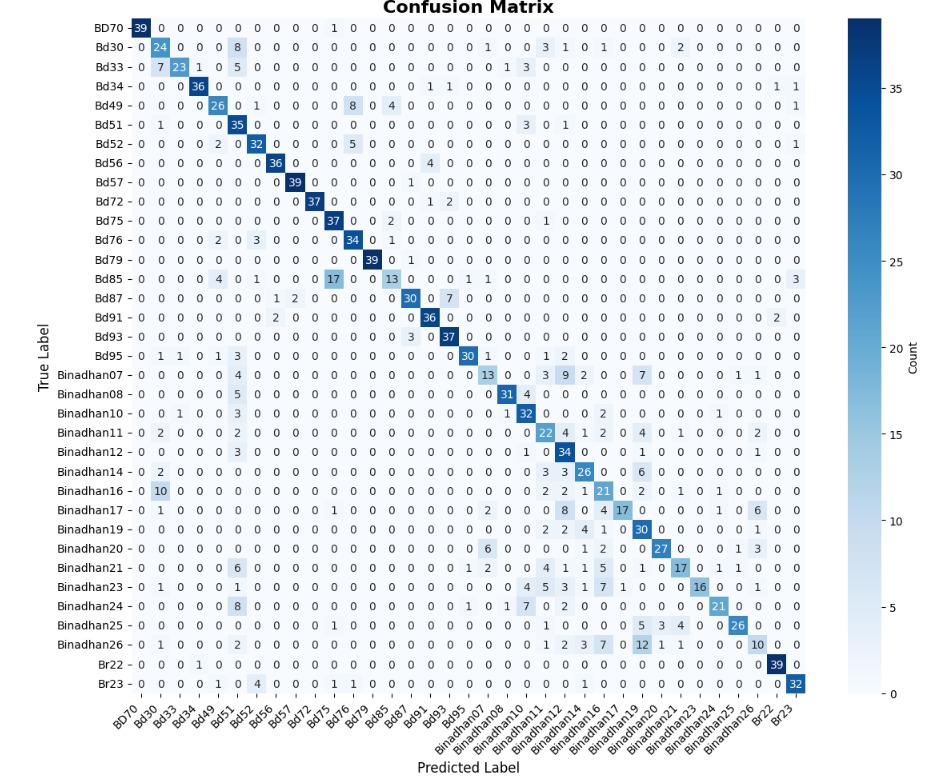}
    \par\vspace{1ex}(b) Deeper Baseline Model (With Bottleneck)
\end{minipage}

\caption{Confusion matrices of the two models on the PaddyVarietyBD Dataset (Test Set).}
\label{fig:paddy_confusions}
\end{figure}

We now present the performance of both models in Table~\ref{tab:paddy_results}. We also present the confusion matrices of both models in Figure~\ref{fig:paddy_confusions}, which illustrate their class-wise prediction behavior on the PaddyVarietyBD test set.

\begin{table}[!t]
\caption{Performance on PaddyVarietyBD Dataset (Test Set)}
\label{tab:paddy_results}
\centering
\begin{tabular}{@{}lcccc@{}}
\toprule
\textbf{Model} & \textbf{Accuracy} & \textbf{Precision} & \textbf{Recall} & \textbf{F1-Score} \\
\midrule
Baseline Model (No Bottleneck) & 0.43 & 0.50 & 0.43 & 0.40 \\
Deeper Baseline Model (With Bottleneck) & 0.71 & 0.75 & 0.71 & 0.71 \\
\bottomrule
\end{tabular}
\end{table}

\subsubsection{Detection and Recognition: Auto-RickshawImageBD}

We evaluate the performance of the proposed MiniYOLO model on the Auto-RickshawImageBD dataset to assess its effectiveness in object-level detection and recognition. Table~\ref{tab:miniyolo_results} reports the classification performance of MiniYOLO in terms of accuracy, precision, recall, and F1-score on the test set.

\begin{figure}[!t]
\centering
\includegraphics[width=0.65\linewidth]{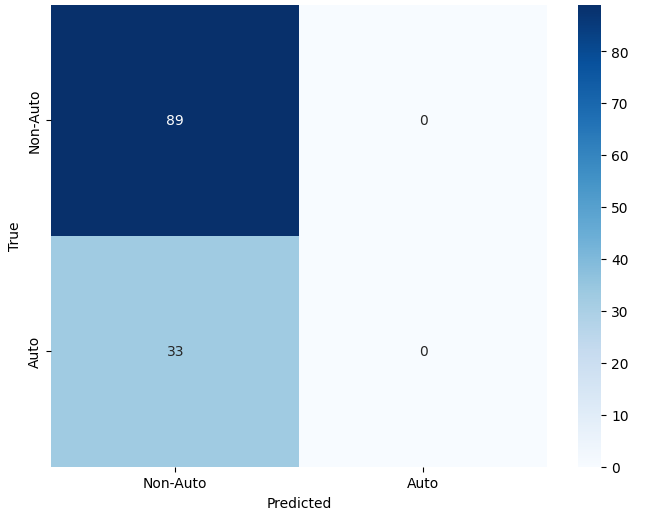}
\caption{Confusion matrix of the MiniYOLO model on the Auto-RickshawImageBD test set.}
\label{fig:miniyolo_confusion}
\end{figure}

The results indicate that MiniYOLO achieves an overall accuracy of 0.73. However, the confusion matrix shown in Figure~\ref{fig:miniyolo_confusion} reveals a strong bias toward the majority class (Non-Auto), with limited recognition of the Auto-Rickshaw class. This behavior reflects the challenges posed by class imbalance and the simplified detection design of the MiniYOLO architecture. Despite these limitations, the model serves as a lightweight baseline for evaluating object-level recognition performance and provides a reference point for comparison with more advanced detection frameworks.

\begin{table}[!t]
\caption{Performance of MiniYOLO on Auto-RickshawImageBD Dataset (Test Set)}
\label{tab:miniyolo_results}
\centering
\begin{tabular}{@{}lcccc@{}}
\toprule
\textbf{Model} & \textbf{Accuracy} & \textbf{Precision} & \textbf{Recall} & \textbf{F1-Score} \\
\midrule
MiniYOLO & 0.73 & 0.53 & 0.73 & 0.62 \\
\bottomrule
\end{tabular}
\end{table}

\subsection{External Model Comparisons: Pretrained and Transfer Learning}

We now compare the best-performing custom CNN architectures against pretrained and transfer learning models under identical evaluation settings for each dataset.

\subsubsection{Binary Classification: Road Damage Dataset}

Table~\ref{tab:road_damage_external} presents a comparative evaluation of the best-performing custom CNN architecture (Evolved Baseline) against pretrained and transfer learning variants of MobileNetV2 and EfficientNetB0 on the Road Damage dataset. All models were evaluated using identical data splits and consistent metrics, including accuracy, precision, recall, F1-score, training time per epoch, and parameter footprint.

\begin{table}[!t]
\caption{External Model Comparison on Road Damage Dataset (Test Set)}
\label{tab:road_damage_external}
\centering
\setlength{\tabcolsep}{3pt} % default is 6pt
\begin{tabular}{@{}lccccccc@{}}
\toprule
\textbf{Model} & \textbf{Acc.} & \textbf{Prec.} & \textbf{Rec.} & \textbf{F1} &
\makecell{\textbf{Time /}\\\textbf{Epoch (s)}} &
\makecell{\textbf{Trainable}\\\textbf{(MB)}} &
\makecell{\textbf{Frozen}\\\textbf{(MB)}} \\
\midrule
Evolved Baseline (Custom CNN) 
& 0.78 & 0.77 & 0.78 & 0.75 & 60.00 & 42.64 & 0.04 \\

MobileNetV2 (Pretrained)     
& 0.72 & 0.88 & 0.72 & 0.74 & 0.90 & 8.49 & 0.13 \\

MobileNetV2 (Transfer)       
& 0.91 & 0.90 & 0.91 & 0.90 & 0.24 & 0.01 & 8.61 \\

EfficientNetB0 (Pretrained)  
& 0.94 & 0.94 & 0.94 & 0.94 & 1.25 & 15.29 & 0.16 \\

EfficientNetB0 (Transfer)    
& 0.78 & 0.61 & 0.78 & 0.69 & 0.33 & 0.01 & 15.45 \\
\bottomrule
\end{tabular}
\end{table}

Among all evaluated models, the \textbf{pretrained EfficientNetB0} achieved the strongest overall performance, with an accuracy, precision, recall, and F1-score of \textbf{0.94}. This highlights the benefit of full fine-tuning, which allows the pretrained backbone to adapt effectively to the Road Damage task—albeit at a moderate cost of \textbf{1.25 seconds per epoch}.

The \textbf{transfer learning version of MobileNetV2} also performed strongly, achieving an F1-score of \textbf{0.90} with only \textbf{0.24 seconds per epoch} of training time and minimal parameter updates. This makes it a favorable choice for resource-constrained deployment scenarios.

In contrast, the \textbf{Evolved Baseline custom CNN} achieved competitive performance (F1-score \textbf{0.75}) but with a substantially higher computational cost—over \textbf{42 MB} of trainable parameters and \textbf{60 seconds per epoch} of training time. This underscores the efficiency advantage of pretrained models.

The \textbf{pretrained MobileNetV2} delivered moderate results, while the \textbf{transfer learning variant of EfficientNetB0} showed lower precision. This suggests that freezing the backbone can limit adaptation to the target task in certain cases. Overall, the results reveal clear trade-offs between classification accuracy, model size, and training efficiency across different configurations.

\begin{figure}[!t]
\centering

% Row 1
\begin{minipage}[b]{0.48\textwidth}
    \centering
    \includegraphics[width=0.95\linewidth]{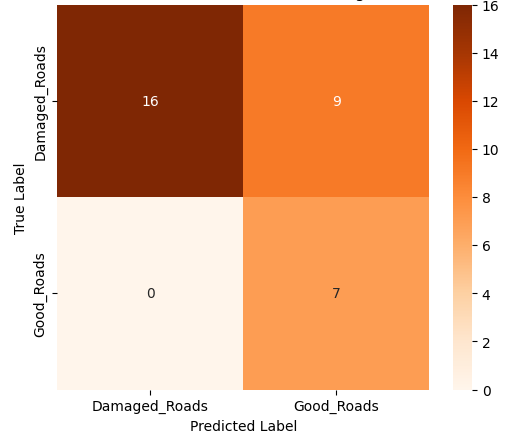}
    \par\vspace{1ex}(a) MobileNetV2 (Pretrained)
\end{minipage}
\hfill
\begin{minipage}[b]{0.48\textwidth}
    \centering
    \includegraphics[width=0.95\linewidth]{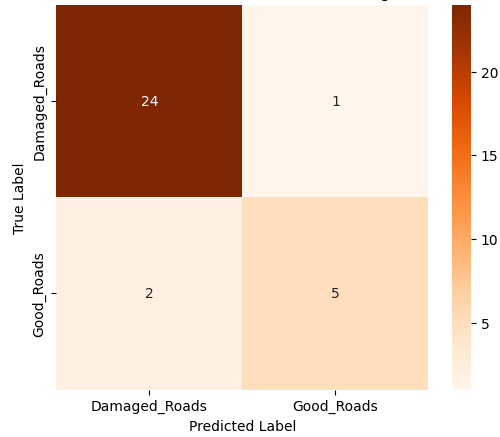}
    \par\vspace{1ex}(b) MobileNetV2 (Transfer Learning)
\end{minipage}

\vspace{0.5cm}

% Row 2
\begin{minipage}[b]{0.48\textwidth}
    \centering
    \includegraphics[width=0.95\linewidth]{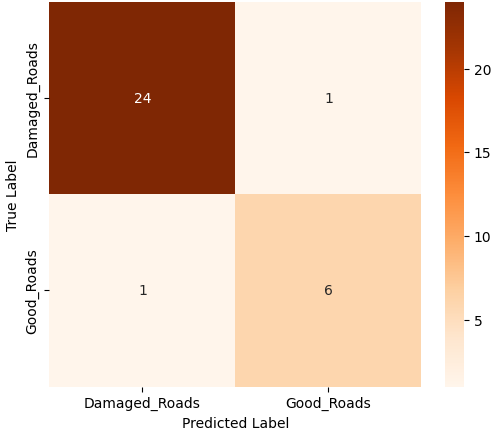}
    \par\vspace{1ex}(c) EfficientNetB0 (Pretrained)
\end{minipage}
\hfill
\begin{minipage}[b]{0.48\textwidth}
    \centering
    \includegraphics[width=0.95\linewidth]{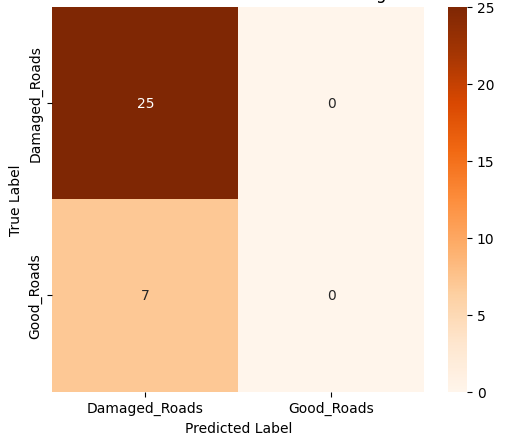}
    \par\vspace{1ex}(d) EfficientNetB0 (Transfer Learning)
\end{minipage}

\caption{Confusion matrices of four pretrained and transfer learning variants on the Road Damage Dataset (Test Set).}
\label{fig:road_damage_external_confusions}
\end{figure}

The evolved custom baseline confusion matrix is already presented in Figure~\ref{fig:road_damage_confusions}. Figure~\ref{fig:road_damage_external_confusions} now presents the confusion matrices for all four pretrained and transfer learning variants, illustrating their class-wise prediction behavior on the Road Damage test set.  Notably, EfficientNetB0 (Transfer Learning) exhibited perfect recall for the majority "Damaged" class but failed to predict any "Good" class instances, highlighting the impact of class imbalance on fixed-feature models. Interestingly, MobileNetV2 (Transfer Learning) achieved a class-wise prediction distribution that closely resembles the much heavier pretrained EfficientNetB0, demonstrating its robustness despite a lightweight architecture.

\subsubsection{Binary Classification: Footpath Dataset}

Table~\ref{tab:footpath_external} presents the comparative performance of the best-performing custom CNN architecture (Evolved Baseline) alongside pretrained and transfer learning variants of MobileNetV2 and EfficientNetB0 on the Footpath dataset. All models were evaluated using identical data splits and consistent evaluation metrics, including accuracy, precision, recall, F1-score, training time per epoch, and parameter footprint.

\begin{table}[!t]
\caption{External Model Comparison on Footpath Dataset (Test Set)}
\label{tab:footpath_external}
\centering
\setlength{\tabcolsep}{3pt} % default is 6pt
\begin{tabular}{@{}lccccccc@{}}
\toprule
\textbf{Model} & \textbf{Acc.} & \textbf{Prec.} & \textbf{Rec.} & \textbf{F1} &
\makecell{\textbf{Time /}\\\textbf{Epoch (s)}} &
\makecell{\textbf{Trainable}\\\textbf{(MB)}} &
\makecell{\textbf{Frozen}\\\textbf{(MB)}} \\
\midrule
Evolved Baseline (Custom CNN) 
& 0.71 & 0.71 & 0.71 & 0.71 & 126.00 & 42.64 & 0.04 \\

MobileNetV2 (Pretrained)     
& 0.79 & 0.85 & 0.79 & 0.79 & 2.26 & 8.49 & 0.13 \\

MobileNetV2 (Transfer)       
& 0.89 & 0.89 & 0.89 & 0.89 & 0.59 & 0.01 & 8.61 \\

EfficientNetB0 (Pretrained)  
& 0.75 & 0.76 & 0.75 & 0.75 & 3.29 & 15.29 & 0.16 \\

EfficientNetB0 (Transfer)    
& 0.63 & 0.62 & 0.63 & 0.62 & 0.81 & 0.01 & 15.45 \\
\bottomrule
\end{tabular}
\end{table}

\begin{figure}[!t]
\centering

% Row 1
\begin{minipage}[b]{0.48\textwidth}
    \centering
    \includegraphics[width=0.95\linewidth]{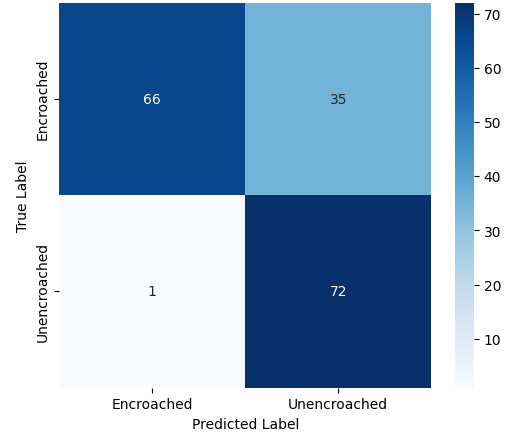}
    \par\vspace{1ex}(a) MobileNetV2 (Pretrained)
\end{minipage}
\hfill
\begin{minipage}[b]{0.48\textwidth}
    \centering
    \includegraphics[width=0.95\linewidth]{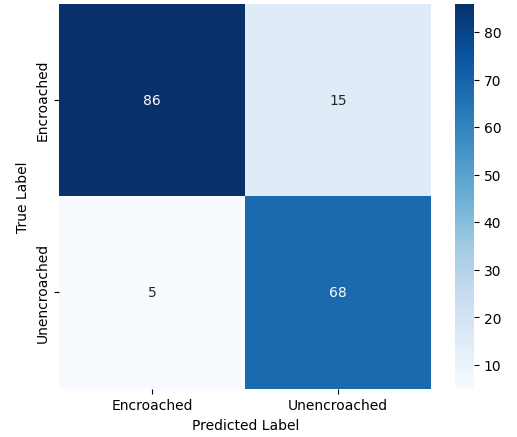}
    \par\vspace{1ex}(b) MobileNetV2 (Transfer Learning)
\end{minipage}

\vspace{0.5cm}

% Row 2
\begin{minipage}[b]{0.48\textwidth}
    \centering
    \includegraphics[width=0.95\linewidth]{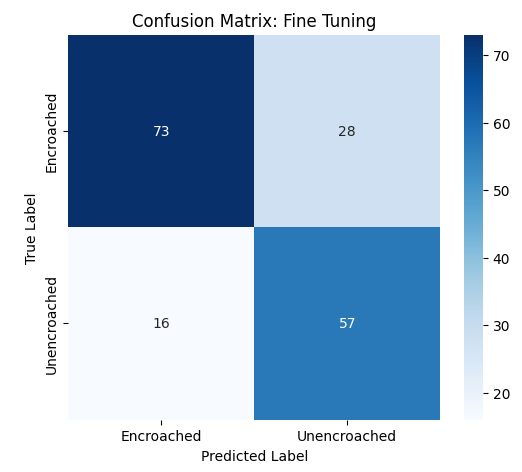}
    \par\vspace{1ex}(c) EfficientNetB0 (Pretrained)
\end{minipage}
\hfill
\begin{minipage}[b]{0.48\textwidth}
    \centering
    \includegraphics[width=0.95\linewidth]{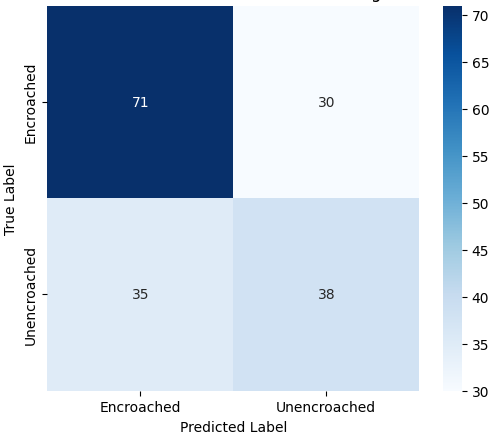}
    \par\vspace{1ex}(d) EfficientNetB0 (Transfer Learning)
\end{minipage}

\caption{Confusion matrices of four pretrained and transfer learning variants on the Footpath Dataset (Test Set).}
\label{fig:footpath_external_confusions}
\end{figure}

Among the evaluated models, the \textbf{transfer learning variant of MobileNetV2} achieved the strongest overall performance on the Footpath dataset, attaining an accuracy, precision, recall, and F1-score of \textbf{0.89}. This result was obtained with a relatively low training cost of \textbf{0.59 seconds per epoch} and a minimal number of trainable parameters, making it particularly well-suited for efficient deployment.

The \textbf{pretrained MobileNetV2} also performed competitively, achieving an F1-score of \textbf{0.79}, albeit with a higher training time of \textbf{2.26 seconds per epoch}. In comparison, the \textbf{pretrained EfficientNetB0} exhibited moderate performance (F1-score \textbf{0.75}) while incurring the highest training cost among the pretrained variants.

The \textbf{Evolved Baseline custom CNN} achieved balanced classification performance with an F1-score of \textbf{0.71}, but required substantially greater computational resources, with over \textbf{42 MB} of trainable parameters and a training time of approximately \textbf{126 seconds per epoch}. This highlights the overhead associated with training deeper custom architectures from scratch.

Finally, the \textbf{transfer learning variant of EfficientNetB0} demonstrated the weakest performance on this dataset, suggesting that freezing the backbone limited its ability to adapt to the structural patterns present in Footpath imagery. Overall, these results indicate that lightweight architectures combined with transfer learning can outperform deeper custom and pretrained models on binary classification tasks, while offering significant gains in training efficiency.

The evolved custom baseline confusion matrix for the Footpath dataset is presented earlier in Figure~\ref{fig:footpath_confusions}. Figure~\ref{fig:footpath_external_confusions} illustrates the confusion matrices for the four pretrained and transfer learning variants, offering insight into their class-wise prediction behavior on the Footpath test set. Notably, the transfer learning variant of MobileNetV2 demonstrates a well-balanced distribution between the \textit{Encroached} and \textit{Unencroached} classes, consistent with its strong overall accuracy and F1-score. In contrast, EfficientNetB0 under transfer learning exhibits a pronounced bias toward the majority class, resulting in reduced recall for the minority category. The pretrained variants also show elevated misclassification rates for encroached regions. Overall, these findings underscore that lightweight architectures—when selectively fine-tuned—can achieve more balanced class-wise predictions.

\subsubsection{Multiclass Classification: MangoImageBD Dataset}

Table~\ref{tab:mango_external} presents a comparative evaluation of the enhanced custom baseline model against pretrained and transfer learning variants of MobileNetV2 and EfficientNetB0 on the MangoImageBD dataset. All models were evaluated under consistent experimental settings using accuracy, precision, recall, F1-score, training time per epoch, and parameter footprint.

\begin{table}[!t]
\caption{External Model Comparison on MangoImageBD Dataset (Test Set)}
\label{tab:mango_external}
\centering
\setlength{\tabcolsep}{2pt} % default is 6pt
\begin{tabular}{@{}lccccccc@{}}
\toprule
\textbf{Model} & \textbf{Acc.} & \textbf{Prec.} & \textbf{Rec.} & \textbf{F1} &
\makecell{\textbf{Time /}\\\textbf{Epoch (s)}} &
\makecell{\textbf{Trainable}\\\textbf{(MB)}} &
\makecell{\textbf{Frozen}\\\textbf{(MB)}} \\
\midrule
Enhanced Baseline (Custom CNN) 
& 0.98 & 0.98 & 0.98 & 0.98 & 36.00 & 89.79 & 0.21 \\

MobileNetV2 (Pretrained)     
& 0.58 & 0.52 & 0.58 & 0.52 & 21.00 & 8.56 & 0.13 \\

MobileNetV2 (Transfer)       
& 0.92 & 0.92 & 0.92 & 0.92 & 7.92 & 0.07 & 8.61 \\

EfficientNetB0 (Pretrained)  
& 0.95 & 0.95 & 0.95 & 0.95 & 27.00 & 15.36 & 0.16 \\

EfficientNetB0 (Transfer)    
& 0.93 & 0.93 & 0.93 & 0.93 & 9.72 & 0.07 & 15.45 \\
\bottomrule
\end{tabular}
\end{table}

The \textbf{enhanced custom baseline model} achieved the highest overall performance on MangoImageBD, reaching an accuracy and F1-score of \textbf{0.98}. This underscores the benefit of deeper architectures with bottleneck residual blocks in fine-grained multiclass recognition, where subtle visual distinctions demand greater representational power. Though this comes at the cost of increased training time and a larger parameter footprint, the performance gains justify the trade-off in high-accuracy settings.

Among the pretrained models, \textbf{EfficientNetB0 (Pretrained)} performed strongly with an F1-score of \textbf{0.95}, albeit at a higher computational cost than its transfer learning variant. In contrast, the \textbf{pretrained MobileNetV2} struggled, suggesting that shallower architectures are less suited for fine-grained tasks.

Both transfer learning models offered a compelling balance between efficiency and accuracy. \textbf{EfficientNetB0 (Transfer)} achieved an F1-score of \textbf{0.93} with minimal trainable parameters, while \textbf{MobileNetV2 (Transfer)} also delivered competitive performance at lower training cost. Overall, the results suggest that deeper custom networks excel in complex multiclass tasks, while transfer learning provides a practical trade-off between efficiency and performance.

\begin{figure}[H]
\centering

% Row 1
\begin{minipage}[b]{0.8\textwidth}
    \centering
    \includegraphics[width=\linewidth]{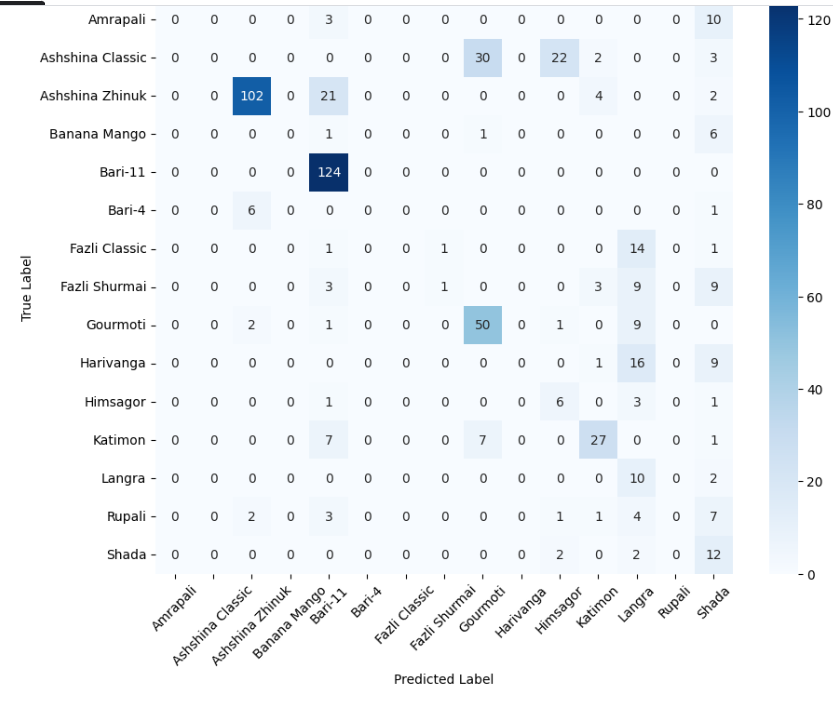}
    \par\vspace{1ex}(a) MobileNetV2 (Pretrained)
\end{minipage}

\vspace{0.60cm}

% Row 2
\begin{minipage}[b]{0.8\textwidth}
    \centering
    \includegraphics[width=\linewidth]{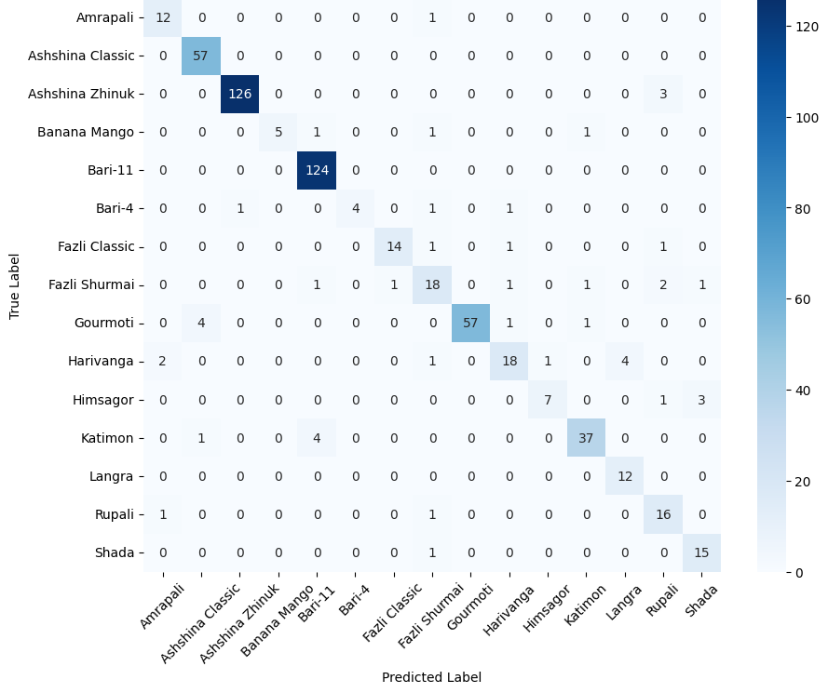}
    \par\vspace{1ex}(b) MobileNetV2 (Transfer Learning)
\end{minipage}

\caption{Confusion matrices of MobileNetV2 variants on the MangoImageBD Dataset (Test Set).}
\label{fig:mango_mobilenet_confusions}
\end{figure}

\begin{figure}[H]
\centering

% Row 1
\begin{minipage}[b]{0.8\textwidth}
    \centering
    \includegraphics[width=\linewidth]{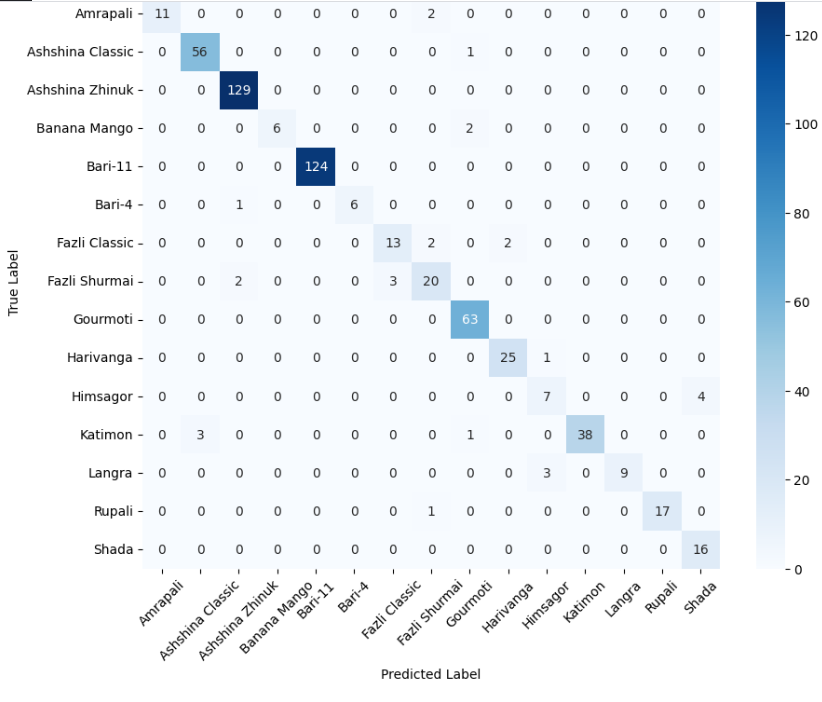}
    \par\vspace{1ex}(a) EfficientNetB0 (Pretrained)
\end{minipage}

\vspace{0.60cm}

% Row 2
\begin{minipage}[b]{0.8\textwidth}
    \centering
    \includegraphics[width=\linewidth]{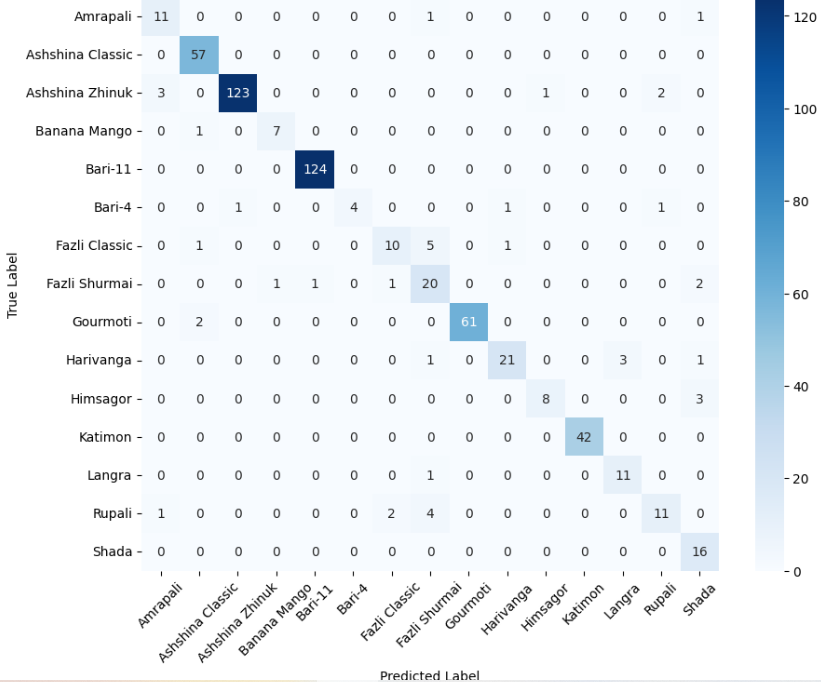}
    \par\vspace{1ex}(b) EfficientNetB0 (Transfer Learning)
\end{minipage}

\caption{Confusion matrices of EfficientNetB0 variants on the MangoImageBD Dataset (Test Set).}
\label{fig:mango_efficientnet_confusions}
\end{figure}

The confusion matrix of the enhanced custom baseline model is shown earlier in Figure~\ref{fig:mango_confusions}. Figures~\ref{fig:mango_mobilenet_confusions} and~\ref{fig:mango_efficientnet_confusions} present the confusion matrices for the MobileNetV2 and EfficientNetB0 variants, respectively, under both pretrained and transfer learning settings. Notably, the pretrained EfficientNetB0 exhibits strong diagonal dominance across most mango categories, indicating robust class-wise discrimination for fine-grained recognition. The transfer learning variant of EfficientNetB0 maintains a similar pattern with slightly increased off-diagonal confusion, suggesting limited loss in performance despite freezing the backbone. In contrast, MobileNetV2 variants show higher confusion among visually similar classes, particularly in the pretrained setting, reflecting the challenges faced by lighter architectures in fine-grained multiclass tasks. Overall, these confusion patterns are consistent with the quantitative results and highlight the role of model depth and feature capacity in distinguishing closely related mango varieties.

\subsubsection{Multiclass Classification: PaddyVarietyBD Dataset}

Table~\ref{tab:paddy_external} presents a comparative evaluation of the enhanced custom CNN baseline against pretrained and transfer learning variants of MobileNetV2 and EfficientNetB0 on the PaddyVarietyBD dataset. All models were evaluated under identical experimental settings using accuracy, precision, recall, F1-score, training time per epoch, and parameter footprint.

\begin{table}[!t]
\caption{External Model Comparison on PaddyVarietyBD Dataset (Test Set)}
\label{tab:paddy_external}
\centering
\setlength{\tabcolsep}{3pt} % default is 6pt
\begin{tabular}{@{}lccccccc@{}}
\toprule
\textbf{Model} & \textbf{Acc.} & \textbf{Prec.} & \textbf{Rec.} & \textbf{F1} &
\makecell{\textbf{Time /}\\\textbf{Epoch (s)}} &
\makecell{\textbf{Trainable}\\\textbf{(MB)}} &
\makecell{\textbf{Frozen}\\\textbf{(MB)}} \\
\midrule
Enhanced Baseline (Custom CNN) 
& 0.71 & 0.75 & 0.71 & 0.71 & 88.00 & 89.95 & 0.21 \\

MobileNetV2 (Pretrained)     
& 0.61 & 0.72 & 0.61 & 0.58 & 51.48 & 8.65 & 0.13 \\

MobileNetV2 (Transfer)       
& 0.56 & 0.65 & 0.56 & 0.55 & 12.58 & 0.17 & 8.61 \\

EfficientNetB0 (Pretrained)  
& 0.70 & 0.79 & 0.70 & 0.69 & 79.20 & 15.46 & 0.16 \\

EfficientNetB0 (Transfer)    
& 0.44 & 0.69 & 0.44 & 0.42 & 15.66 & 0.17 & 15.45 \\
\bottomrule
\end{tabular}
\end{table}

The \textbf{enhanced custom CNN baseline} achieved the strongest overall performance on the PaddyVarietyBD dataset, with an accuracy and F1-score of \textbf{0.71}. This indicates that increased depth and bottleneck residual blocks are effective for capturing the subtle inter-class variations present in fine-grained paddy grain imagery—\textit{though at the cost of significantly higher training time and model size}.

Among the pretrained models, \textbf{EfficientNetB0 (Pretrained)} performed competitively, achieving an F1-score of \textbf{0.69}, while the \textbf{pretrained MobileNetV2} showed comparatively lower performance.

Both transfer learning variants exhibited reduced performance on this dataset. In particular, \textbf{EfficientNetB0 (Transfer Learning)} showed a notable drop in recall, indicating limited adaptability when the backbone is frozen. Although transfer learning significantly reduced training time and the number of trainable parameters, these results suggest that fine-grained multiclass tasks such as PaddyVarietyBD benefit from full feature adaptation or deeper custom architectures. Overall, the findings highlight a trade-off between computational efficiency and classification performance for highly granular visual recognition problems.

\begin{figure}[H]
\centering

% Row 1
\begin{minipage}[b]{0.8\textwidth}
    \centering
    \includegraphics[width=\linewidth]{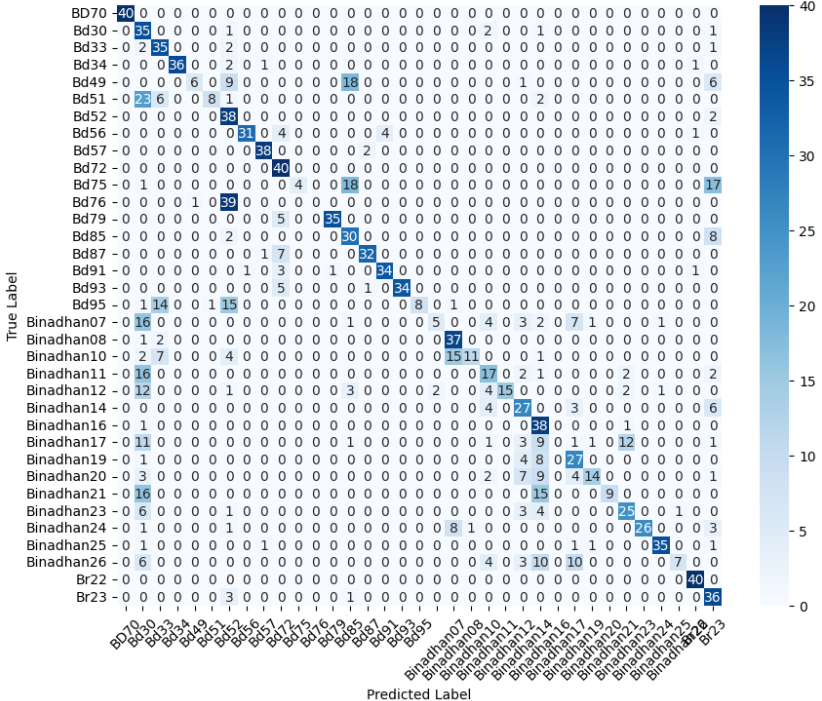}
    \par\vspace{1ex}(a) MobileNetV2 (Pretrained)
\end{minipage}

\vspace{0.60cm}

% Row 2
\begin{minipage}[b]{0.8\textwidth}
    \centering
    \includegraphics[width=\linewidth]{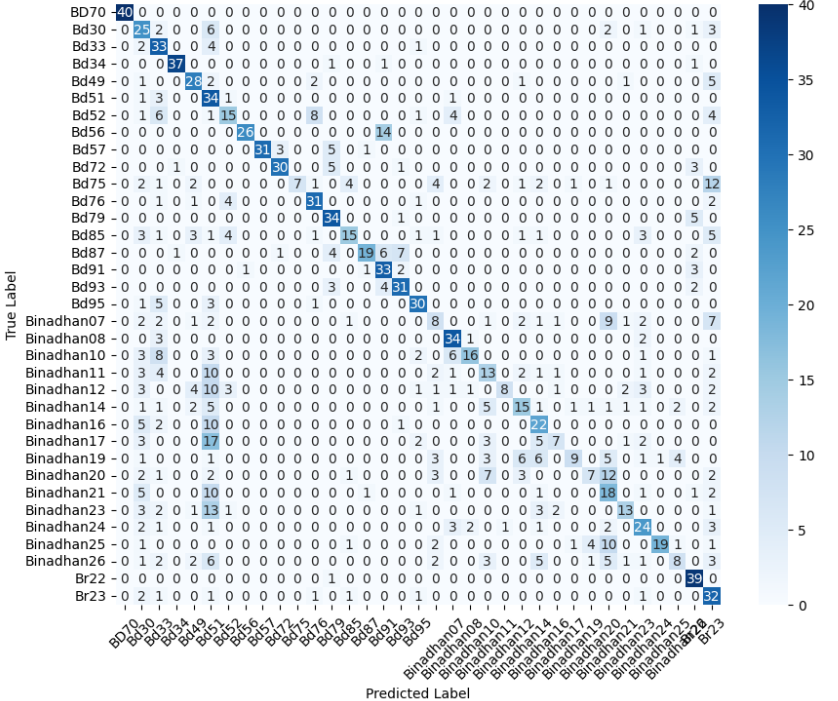}
    \par\vspace{1ex}(b) MobileNetV2 (Transfer Learning)
\end{minipage}

\caption{Confusion matrices of MobileNetV2 variants on the PaddyVarietyBD Dataset (Test Set).}
\label{fig:paddy_mobilenet_confusions}
\end{figure}

\begin{figure}[H]
\centering

% Row 1
\begin{minipage}[b]{0.8\textwidth}
    \centering
    \includegraphics[width=\linewidth]{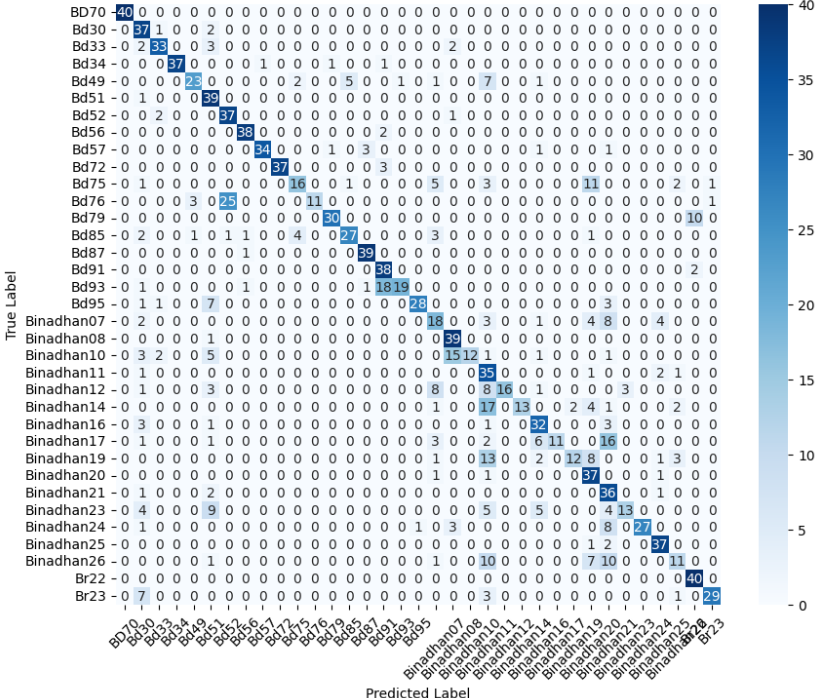}
    \par\vspace{1ex}(a) EfficientNetB0 (Pretrained)
\end{minipage}

\vspace{0.60cm}

% Row 2
\begin{minipage}[b]{0.8\textwidth}
    \centering
    \includegraphics[width=\linewidth]{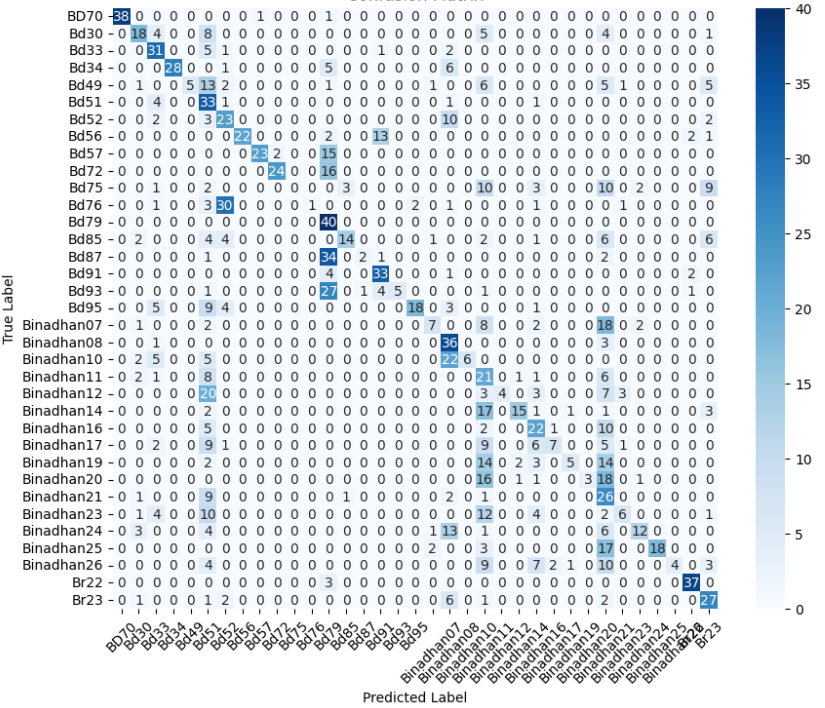}
    \par\vspace{1ex}(b) EfficientNetB0 (Transfer Learning)
\end{minipage}

\caption{Confusion matrices of EfficientNetB0 variants on the PaddyVarietyBD Dataset (Test Set).}
\label{fig:paddy_efficientnet_confusions}
\end{figure}

The confusion matrix of the enhanced custom baseline model for the PaddyVarietyBD dataset is presented earlier in Figure~\ref{fig:paddy_confusions}. Figures~\ref{fig:paddy_mobilenet_confusions} and~\ref{fig:paddy_efficientnet_confusions} show the confusion matrices for the MobileNetV2 and EfficientNetB0 variants, respectively, under pretrained and transfer learning settings. Across both architectures, the pretrained variants exhibit stronger diagonal dominance than their transfer learning counterparts, indicating improved class-wise discrimination when all network layers are trainable. In contrast, the transfer learning models show increased off-diagonal confusion, particularly among visually similar paddy varieties, reflecting limited adaptability when the backbone is frozen. While EfficientNetB0 demonstrates relatively more stable predictions than MobileNetV2, both lightweight transfer learning models struggle to separate closely related grain types. These confusion patterns align with the quantitative results and highlight the importance of deeper feature adaptation for fine-grained agricultural classification tasks.

\subsubsection{Detection and Recognition: Auto-RickshawImageBD Dataset}

Table~\ref{tab:autorickshaw_detection_external} presents a comparative evaluation of the MiniYOLO model alongside pretrained and transfer learning variants of Faster R-CNN and YOLO on the Auto-RickshawImageBD dataset. All models were assessed using consistent metrics—accuracy, precision, recall, F1-score—as well as training time per epoch and model size (parameter count).

\begin{table}[t!]
\caption{Comparison of Detection Models on Auto-RickshawImageBD Dataset (Test Set)}
\label{tab:autorickshaw_detection_external}
\centering
\begin{tabular}{@{}lcccccc@{}}
\toprule
\textbf{Model} & \textbf{Acc.} & \textbf{Prec.} & \textbf{Rec.} & \textbf{F1} & \textbf{Time/Epoch (s)} & \textbf{Params} \\ 
\midrule
MiniYOLO (Custom) 
& 0.73 & 0.53 & 0.73 & 0.62 & 80.0 & 98.2K \\

Faster R-CNN (Pretrained) 
& 0.76 & 0.93 & 0.76 & 0.76 & 300.0 & 41.3M \\

Faster R-CNN (Transfer Learning) 
& 0.00 & 0.00 & 0.00 & 0.00 & 73.7 & 41.3M \\

YOLO (Pretrained) 
& 0.56 & 0.57 & 0.56 & 0.56 & 50.0 & 3.0M \\

YOLO (Transfer Learning) 
& 0.31 & 0.41 & 0.25 & 0.31 & 35.9 & 3.0M \\
\bottomrule
\end{tabular}
\end{table}

Among all evaluated models, the \textbf{pretrained Faster R-CNN} achieved the best detection performance, with an F1-score of \textbf{0.76} and high precision. This suggests effective localization and classification when the full pretrained backbone is fine-tuned to the task, albeit with a significant computational cost (around \textbf{300 seconds per epoch}) and large model size.

The \textbf{MiniYOLO} model performed competitively, reaching an accuracy of \textbf{0.73} and F1-score of \textbf{0.62} with only \textbf{98.2K parameters}. This demonstrates its potential as a lightweight baseline for detection tasks in resource-constrained environments.

In contrast, the \textbf{transfer learning variant of Faster R-CNN} failed to detect objects altogether, yielding zero accuracy and F1-score. This is due to a zero IoU score during evaluation, indicating that the frozen backbone could not capture the dataset’s spatial patterns, resulting in failed localization.

For the YOLO-based models, the \textbf{pretrained YOLO} outperformed its transfer learning counterpart, achieving an F1-score of \textbf{0.56}. The \textbf{transfer learning YOLO} variant exhibited limited recall and generalization, reinforcing the need for full fine-tuning to adapt to the visual characteristics of auto-rickshaws in dense traffic scenes.

\section{Discussion}

This section synthesizes the experimental findings across all evaluated datasets, focusing on how architectural design choices and training strategies influence classification performance, computational efficiency, and model scalability. We first analyze insights gained from progressively evolving custom CNN architectures, followed by a detailed examination of the trade-offs between custom, pretrained, and transfer learning models under varying task complexities.

\subsection{Architectural Insights from Custom CNN Experiments}

The experiments conducted across five datasets reveal several consistent trends regarding the behavior of the custom CNN and its evolved variants. For the two binary classification tasks—Road Damage and FootpathVision—the Evolved Baseline model achieved the most stable and balanced performance. Its initial 7$\times$7 convolution and simplified classification head enabled better generalization compared to the custom architecture, which struggled with class imbalance and frequently misclassified the minority class.

In contrast, the fine-grained multiclass datasets demanded significantly greater representational capacity. Both MangoImageBD and PaddyVarietyBD showed marked improvements only when the deeper baseline model with bottleneck residual blocks was employed. The baseline without bottlenecks lacked the depth needed to distinguish subtle inter-class variations, resulting in low accuracy and weak F1-scores. Once bottleneck blocks were introduced, the model captured richer hierarchical features, leading to substantial accuracy gains—reaching 0.98 for mango varieties and 0.71 for paddy grains. These results highlight the importance of architectural depth and channel expansion in high-resolution, fine-grained visual recognition tasks.

The findings also underscore a clear divide between the requirements of binary and multiclass classification. While lighter architectures are sufficient for tasks involving coarse visual distinctions, fine-grained datasets consistently benefit from deeper residual models capable of learning more abstract representations. Thus, as classification complexity increases, the role of residual pathways and enhanced feature dimensionality becomes increasingly critical.

Finally, the Auto-RickshawImageBD experiment demonstrates that classification-oriented backbones can be effectively adapted for region-aware detection when paired with a proposal-based prediction head. Despite being evaluated on only 8 images, the detector exhibited strong localization performance, achieving an average IoU of 0.7366 along with high precision. This suggests that the feature extraction strategies explored in earlier classification experiments transfer well to object-level tasks when integrated into a detection pipeline.

Overall, this study highlights that no single CNN architecture is universally optimal. Instead, architectural suitability depends heavily on dataset characteristics: lightweight models perform well in binary settings with coarse visual cues, whereas deeper bottleneck-based architectures are essential for fine-grained multiclass recognition. The detection results further show that, with appropriate adaptation, these classification backbones can be extended effectively to spatial localization tasks.

\subsection{Trade-Off Analysis Between Custom, Pretrained, and Transfer Learning Models}

Across all evaluated datasets, a clear set of trade-offs emerges between classification performance, computational efficiency, and model complexity. Custom CNN architectures, pretrained models, and transfer learning variants each exhibit distinct strengths and limitations depending on task difficulty and dataset characteristics.

For binary classification tasks such as Road Damage and FootpathVision, lightweight pretrained and transfer learning models consistently offered superior efficiency. In particular, MobileNetV2 under transfer learning achieved strong accuracy and F1-scores while requiring minimal training time and only a small number of trainable parameters. Although the pretrained EfficientNetB0 achieved marginally higher accuracy on the Road Damage dataset, this improvement came at the cost of increased model size and longer training time. At the same time, the evolved custom CNN baseline—although capable of balanced predictions—incurred significantly higher computational costs, with substantially larger parameter footprints and longer training times. This highlights that for coarse-grained binary tasks, pretrained backbones can deliver competitive or superior performance without the overhead of training deep custom architectures from scratch or retraining the entire pretrained backbone.

The trade-offs shift markedly for fine-grained multiclass classification tasks. On MangoImageBD and PaddyVarietyBD, deeper custom CNNs with bottleneck residual blocks achieved the strongest overall performance, particularly in terms of F1-score and class-wise discrimination. These gains came at the expense of increased model size and training time, with trainable parameters approaching \textbf{90 MB} and per-epoch training times reaching approximately \textbf{36 seconds} for MangoImageBD and \textbf{88 seconds} for PaddyVarietyBD. While pretrained EfficientNetB0 models also performed competitively, transfer learning variants generally exhibited reduced recall and increased confusion among visually similar classes, indicating limited adaptability when the backbone is frozen. Thus, efficiency gains from transfer learning often resulted in diminished accuracy for highly granular recognition tasks.

The object detection experiments on the Auto-RickshawImageBD dataset further highlight the sensitivity of spatial localization tasks to training strategy and architectural capacity. While pretrained detection models such as Faster R-CNN demonstrated strong localization and classification performance when fully fine-tuned, freezing the backbone during transfer learning proved ineffective for this dataset, resulting in near-zero IoU and complete detection failure. This indicates that object detection in complex traffic scenes requires full feature adaptation to accurately capture spatial structure and object boundaries. The MiniYOLO model, despite its simplified design and extremely small parameter footprint, achieved reasonable accuracy and recall, serving as a lightweight reference point rather than a high-capacity detector. However, the performance gap between MiniYOLO and fully pretrained detection models underscores the importance of richer spatial feature representations for reliable object-level recognition.

Overall, these findings reinforce that no single CNN architecture is universally optimal. Instead, architectural selection should be guided by dataset complexity, task requirements, and deployment constraints. Lightweight pretrained and transfer learning models are effective for binary classification tasks and efficiency-critical settings, where coarse visual distinctions dominate and computational overhead must be minimized. In contrast, deeper custom architectures with greater representational capacity are better suited for fine-grained multiclass recognition, where subtle inter-class variations require richer feature hierarchies. For object detection tasks, the results further indicate that reliable spatial localization typically necessitates full fine-tuning of pretrained backbones, while lightweight detectors offer limited but useful performance baselines when computational or deployment constraints restrict model complexity. Collectively, these observations underscore the importance of aligning model design and training strategy with task-specific demands rather than adopting a one-size-fits-all solution.

\section{Conclusion}

This study explored the comparative effectiveness of a custom CNN architecture, its evolved variants, and a deep residual baseline with bottleneck blocks across five visual recognition tasks, including binary classification, fine-grained multiclass classification, and object detection. The results demonstrate that model suitability is closely tied to task complexity: lightweight models perform adequately on binary tasks, while deeper architectures are essential for fine-grained recognition. Additionally, classification backbones were found to generalize well when extended to region-aware detection. Building on these observations, the empirical results across all datasets further clarify which modeling approaches are most appropriate for each task when both performance metrics and resource constraints are jointly considered.

For binary classification tasks such as Road Damage and FootpathVision, lightweight pretrained and transfer learning models emerge as the most practical choice. In particular, MobileNetV2 under transfer learning consistently achieved strong accuracy and F1-scores while requiring minimal training time and a very small number of trainable parameters. Although the pretrained EfficientNetB0 attained marginally higher accuracy on the Road Damage dataset, this improvement came at the cost of increased model size and longer training time. Consequently, for coarse-grained binary tasks where deployment efficiency is critical, transfer learning with lightweight backbones offers an optimal balance between performance and computational cost without retraining the entire pretrained backbone.

In contrast, the requirements shift substantially for fine-grained multiclass classification tasks. On MangoImageBD, the enhanced custom CNN with bottleneck residual blocks achieved near-perfect performance, outperforming all pretrained and transfer learning variants. While pretrained EfficientNetB0 also delivered strong results, it remained inferior to the deeper custom architecture, which benefited from greater representational capacity. These gains, however, required increased computational resources, with approximately 90~MB of trainable parameters and training times around 36 seconds per epoch. Thus, for applications prioritizing maximum classification accuracy over efficiency, deeper custom architectures are the most suitable choice. A similar trend was observed for PaddyVarietyBD, where subtle inter-class differences demanded deeper feature hierarchies. The enhanced custom CNN again provided the strongest overall performance, while transfer learning models exhibited reduced recall and increased class confusion. Although transfer learning significantly reduced training time (e.g., 12--16 seconds per epoch), the accompanying loss in accuracy suggests that such efficiency gains may not justify the performance degradation for highly granular agricultural classification tasks. In these scenarios, pretrained models with all layers trainable or deeper custom architectures are better suited despite their higher computational cost.

The object detection experiments on the Auto-RickshawImageBD dataset further reinforce the importance of architectural choice and training strategy for spatially aware tasks. Pretrained Faster R-CNN achieved the strongest overall detection performance, benefiting from full backbone fine-tuning to accurately localize and classify auto-rickshaws in complex traffic scenes, albeit at a substantial computational cost. In contrast, freezing the backbone during transfer learning proved ineffective for Faster R-CNN, resulting in failed localization and zero detection accuracy. The MiniYOLO model, despite its simplified design and extremely small parameter footprint, demonstrated reasonable accuracy and recall, highlighting its viability as a lightweight baseline for detection in resource-constrained settings. However, the performance gap between lightweight and fully pretrained detectors underscores the necessity of rich spatial feature adaptation for reliable object-level recognition.

Overall, the results indicate that no single approach is universally optimal. Lightweight transfer learning models are best suited for binary classification tasks with coarse visual distinctions and strict resource constraints, while deeper custom CNNs or fully trainable pretrained models are more appropriate for fine-grained multiclass problems where classification accuracy and class-wise discrimination are paramount. For object detection tasks, the findings further show that full fine-tuning of pretrained detection architectures is often essential to achieve reliable spatial localization, whereas lightweight detection models provide practical performance--efficiency trade-offs when computational resources are limited.

In future work, we plan to investigate hybrid CNN-transformer architectures, attention mechanisms, and automated neural architecture search (NAS) strategies to further enhance performance and adaptability across diverse visual domains.

\bibliography{sn-bibliography}

\end{document}